\documentclass[sigplan,anonymous=false]{acmart}
\usepackage[symbol]{footmisc}
\AtBeginDocument{%
  \providecommand\BibTeX{{%
    \normalfont B\kern-0.5em{\scshape i\kern-0.25em b}\kern-0.8em\TeX}}}

\setcopyright{acmcopyright}
\copyrightyear{2024}
\acmYear{2024}
\acmDOI{XXXXXXX.XXXXXXX}

\acmConference[SETN '24]{13th Hellenic Conference
  on Artificial Intelligence (SETN 2024)}{September 11-13, 2024}{Piraeus, Greece}
\acmPrice{15.00}
\acmISBN{978-1-4503-XXXX-X/18/06}

\usepackage{hyperref}
\usepackage{graphicx}
\graphicspath{ {./figures/} }

\begin{document}

\title{CycleMix: Mixing Source Domains for Domain Generalization in Style-Dependent Data}

\author{Aristotelis Ballas}
\orcid{0000-0003-1683-8433}
\affiliation{%
  \institution{Department of Informatics and Telematics}
  \institution{Harokopio University}
  \streetaddress{Omirou 9, Tavros}
  \city{Athens}
  \country{Greece}
}
\email{aballas@hua.gr}

\author{Christos Diou}
\orcid{0000-0002-2461-1928}
\affiliation{%
  \institution{Department of Informatics and Telematics}
  \institution{Harokopio University}
  \streetaddress{Omirou 9, Tavros}
  \city{Athens}
  \country{Greece}}
\email{cdiou@hua.gr}

\renewcommand{\shortauthors}{Ballas and Diou.}

\begin{abstract}
As deep learning-based systems have become an integral part of everyday life,  limitations in their generalization ability have begun to emerge. Machine learning algorithms typically rely on the i.i.d. assumption, meaning that their training and validation data are expected to follow the same distribution, which does not necessarily hold in practice. In the case of image classification, one frequent reason that algorithms fail to generalize is that they rely on spurious correlations present in training data, such as associating image styles with target classes. These associations may not be present in the unseen test data, leading to significant degradation of their effectiveness. In this work, we attempt to mitigate this Domain Generalization (DG) problem by training a robust feature extractor which disregards features attributed to image-style but infers based on style-invariant image representations. To achieve this, we train CycleGAN models to learn the different styles present in the training data and randomly mix them together to create samples with novel style attributes to improve generalization. Experimental results on the PACS DG benchmark validate the proposed method\footnote{Code available at: \href{https://github.com/aristotelisballas/cyclemix}{https://github.com/aristotelisballas/cyclemix}.}.

%To validate the efficacy of the proposed method we train 
%a ResNet-50 model on the widely-adopted PACS DG benchmark and evaluate 
%against state-of-the-art data augmentation techniques, where our model is able to surpass them and achieve suprisingly good results and boost the overall performance of the baseline by abound 2\%.
\end{abstract}


%%\ccsdesc[500]{Computer systems organization~Embedded systems}
%%\ccsdesc[300]{Computer systems organization~Redundancy}
%%\ccsdesc{Computer systems organization~Robotics}
%%\ccsdesc[100]{Networks~Network reliability}


%  \includegraphics[width=\textwidth]{Hypercolumn Model.png}
%  \caption{Visualization of the proposed model.}

\maketitle

\section{Introduction}

The past few years have been marked by an explosion in the use of Artificial 
Intelligence systems. Spanning from industry \cite{VERDOUW2021103046}, 
medicine \cite{mckinney2020international}, academia 
\cite{bengio2013representation} and even general public use 
\cite{RAY2023121}, AI systems seem to have established themselves in our daily lives. However, despite their success, widely used and state-of-the-art 
models still fail to exhibit generalizable attributes when evaluated on data
that do not adhere to the i.i.d. assumption. During their training, prominent
neural network architectures, such as deep convolutional neural networks 
(CNNs), often learn to infer based on spurious correlations present in the 
data (e.g. backgrounds, features attributed to the image style, etc.) and not
truly class-representative properties \cite{recht2019imagenet, 10233054}. Domain Generalization \cite{10.1109/TPAMI.2022.3195549} attempts to provide insight into the above issues, by building models which are trained on multiple \textit{source} data domains but are able to generalize to previously unseen data (\textit{target} data domains). 

\begin{figure}[t]
	\centering
	\includegraphics[width=\linewidth]{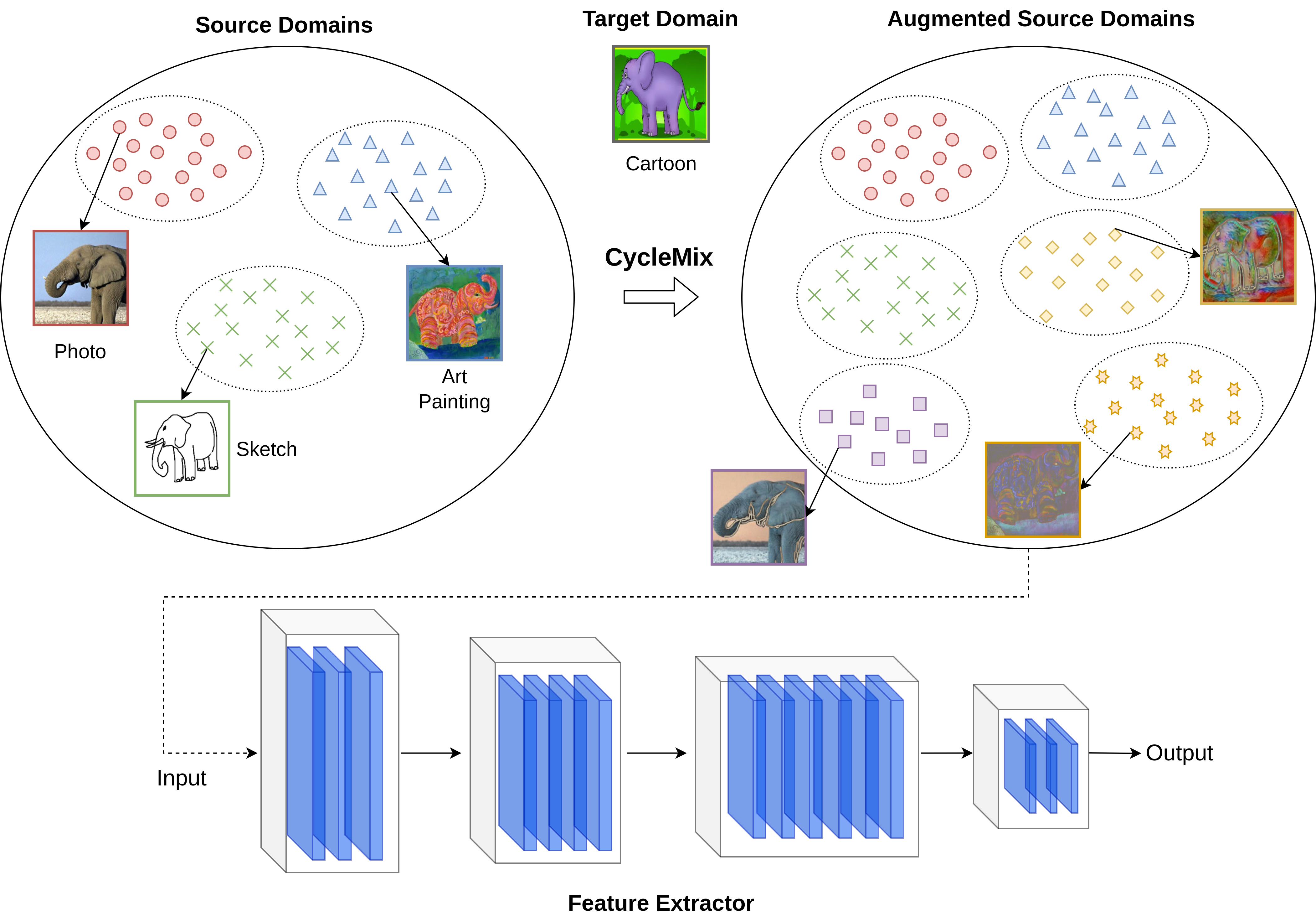}
	\caption{Illustration of the proposed \textit{CycleMix} augmentation method. Before passing through a feature extractor (e.g ResNet-50), the styles of each domain source domain are mixed together in order to create samples from novel domains. As the magniutude of each style component is random in each minibatch, the model is constantly provided with previously unseen data samples during training.}  
	\Description{Our
		proposed framework.}
	\label{fig:cyclemix}
\end{figure}

In our work, we aim to produce a model that maintains its performance on test image data distributions with different \textit{styles} than the ones present in the training distribution. We therefore propose augmenting
the styles of a model's training images and creating novel style image domains which could push a CNN to extract meaningful and domain-invariant representations. Our initial findings are validated on PACS \cite{Li_2017_ICCV}, a widely-used DG benchmark, which contains images from 4 different style domains.

To this end, we:

\begin{itemize}
	\item Train domain translational Generative Adversarial Networks (GANs) for capturing the style attributes of each source domain,
	\item Randomly mix the styles present in the source domains and produce images from novel style domains and
	\item Validate our method on a widely-used publicly available DG dataset.
\end{itemize}

In the sections to come, we briefly: introduce the DG problem setup along with all relevant notations, reference the most important works in DG, 
present the experimental setup and results and finally, conclude our paper.

\subsection{Domain Generalization}

Let $\mathcal{D} := \{\mathcal{D}_i\}_{i=1}^{S}$ a set of $S$ \textit{Source} training domains over an input space $\mathcal{X}$. We then observe $n_i$
training data points from domain $\mathcal{D}_i$, consisting of an input t $\mathbf{x}^{(i)}_j$ and label $y^{(i)}_j$, i.e. $(\mathbf{x}^{(i)}_j, y^{(i)}_j) \sim \mathcal{D}_i$. Similarly, let $T := \{T_i\}_{i=1}^{T}$ be a set of $T$ unknown \textit{Target} domains, while we
assume that there exists a single global labeling function $h(x)$ that maps
input observations to their labels, $y_j^{(i)} = h(\mathbf{x}_j^{(i)})$, for all domains $i$ and samples $j$. The aim of Domain Generalization (DG) is to produce a model with
parameters $\theta \in \Theta$ which generalizes to both source domains $\mathcal{D}$ and unseen target domains $\mathcal{T}$.

%In this section, we formally introduce the notations and definitions of DG.
%Let $X$ be an input (feature) space and $Y$ an output (label) space. A domain is defined
%as a joint distribution $\mathcal{P(X,Y)}$ $\sim$ $\mathcal{P_{XY}}$ on $\mathcal{X \times Y}$ .
%
%In DG, the training and test distributions are OOD, in the sense that we are given S source (training) domains and T target (test) domains,
%where $\mathcal{P}^i_\mathcal{XY}$ $\neq$ $\mathcal{P}^j_\mathcal{XY}$, 1 $\leq$ i, j $\leq$ S, T. Given labeled source domains S, the goal is to learn a model $\mathcal{F}$ ,
%trained on data from S, which can adequately generalize to an unseen domain T.

\section{Related Work}
Domain Generalization has emerged as one of the most difficult problems in ML 
today, finding applications in multiple, varying fields \cite{10.1109/TPAMI.2022.3195549}. DG methods can be 
broadly categorized in the following groups \cite{wang2022generalizing}:
\begin{itemize}
	\item \textbf{Data manipulation}: applied algorithms focus on producing generalizable models by increasing the diversity of existing training data \cite{zhou2021domain}.
	\item \textbf{Representation learning}: given a predictive function $h$, which can be decomposed as $h = f \circ g$ into a representation learning function $g$ and $f$, methods in this group focus on improving the feature extraction capabilities of $g$. The most common approach is regularizing the loss function \cite{arjovsky_invariant_2020} or manipulating existing neural network architectures \cite{10472869, 10.1145/3591106.3592263}.
	\item \textbf{Learning strategy}: the problem of DG has also been studied under numerous machine learning paradigms, such as self-supervised learning, meta-learning, gradient operations, ensemble-learning, etc \cite{ kim_selfreg_2021}. 
\end{itemize}

Our proposed method falls under the first category of data manipulation 
methods. The implementation of data augmentation methods with regards to model generalizabilty has been certainly explored in the past. Most notably the authors of \cite{xu2021robust} explore the benefits of applying random 
convolutions on training images and using them as new data points during 
model training. Several works have also employed adaptive instance normalization (AdaIN) \cite{huang2017arbitrary} for transferring styles 
between data samples. For example, SagNets \cite{nam2021reducing} aim to 
make predictions on the content of an image and disregard features attributed
to image-style by training style-biased networks, while \cite{lee2022bridging} trains a model to learn robust colorization techniques for improved model robustness. In another interesting 
work, the authors of \cite{zhou2021domain} propose MixStyle, an algorithm that
mixes the styles of training instances in each mini-batch to increase the sample diversity of source domains. Another method that produces surprisingly good results is Mixup \cite{xu2020adversarial}, which improves model robustness by training the network on convex combinations of pairs of examples and their labels. Further common augmentation methods or regularization strategies include CutMix \cite{yun2019cutmix} and Cutout \cite{may2020improved}, where patches of images are either cut and pasted among training samples or dropped entirely.

In our work, we propose capturing the style attributes of each domain by employing translational Generative Adversarial Networks and not relying on 
the features present in each separate sample. By mixing the style-attributes of each domain we are able to create completely novel samples in each 
mini-batch and improve the robustness of a vanilla feature extractor.

\section{Methodology}

In this section we provide a brief overview of the CycleGAN algorithm which was utilized in our research for translating images between domains. Once defined, we present the proposed methodology of \textit{CycleMix} for synthesizing images with novel styles .

\subsection{Cycle-Consistent Adversarial Networks}
CycleGANs \cite{zhu2017unpaired} were initially proposed for learning 
translational image mappings between two domains $X$ and $Y$, in the absence 
of paired examples. Formally, let $D_1$ and $D_2$ be source domains for which we aim to learn a mapping $G: D_1 \rightarrow D_2$, such that the 
distribution of images drawn from $G(D_1)$ are identical with the 
distribution of images from $D_2$. The novelty of CycleGAN's lies in the 
addition of an inverse mapping $F: D_1 \rightarrow D_2$ and \textit{Cycle 
Consistency Loss}, to the already established \textit{Adversarial Loss} of 
GAN's. The addition of the Cycle Consistency loss enforces the model to 
reconstruct a translated image into its original domain.

\begin{figure}[t]
	\centering
	\includegraphics[width=0.7\linewidth]{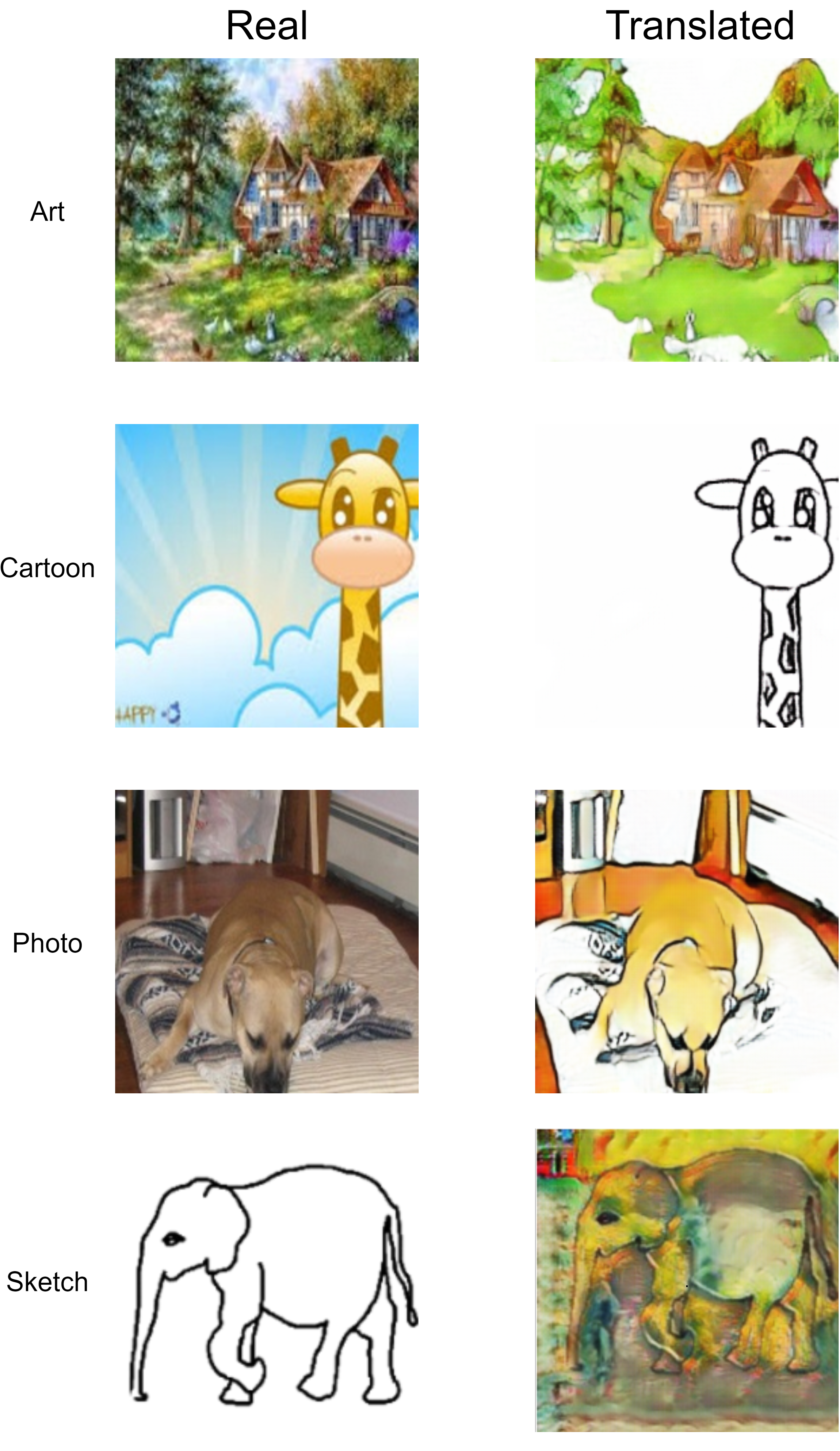}
	\caption{Indicative examples of the trained CycleGAN translations between domains in the PACS dataset. Despite the relatively small size of the samples, the CycleGAN models are able to capture the style attributes in each domain.}  
	\Description{CycleGan}
	\label{fig:cyclegan}
\end{figure}

\subsection{CycleMix}
In our work, we leverage CycleGANs to learn style-mappings between source domains, in the context of Domain Generalization. Our intuition is that by randomly mixing the styles present in source domains and thus synthesizing novel samples, a feature extractor can derive robust representations and learn the features which remain invariant across styles.
%Our intuition is that the style of any target domain can be synthesized by mixing the available styles of the source domains. For example, if the unknown style is that of a \textit{Cartoon}, one can argue that the application of colors found in an \textit{Art} image and bold object outlines present in \textit{Sketches} on a natural image can resemble the style of a \textit{Cartoon}.
Specifically, given $S$ source domains we train $S(S-1)/2$ CycleGANs for learning all the possible domain mappings (a single CycleGAN also contains the trained model yielding the inverse mapping between the two domains). An indicative example of translated images between domains is presented in Fig. \ref{fig:cyclegan}. Having captured the styles of each source domain with the above mappings, we use them as augmentation functions on training data. Specifically, given an image $\mathbf{x}^{(i)}$ drawn from a source domain $D_i$, we translate the image to the remaining source domains and then randomly blend them into a single final image. The entire augmentation operation is as follows:
\begin{equation}
  \mathbf{x}^{(i)'} = \mathbf{x}^{(i)} + \sum_{\substack{j=1 \\ j \neq i}}^{S}a_j \cdot G_{ij}(\mathbf{x}^{(i)})
\end{equation}
where $G_{ij}$ the GAN for translating images from domain $i$ to domain $j$ and $a_j$ a random parameter corresponding to the magnitude of style mixing for each source domain\footnote{The sum of all parameters $a_j$ add up to 1 and are randomly sampled in each minibatch iteration.}. After this mixing operation, each image is normalized (i.e., the same preprocessing is applied as for the vanilla model input).

In our experiments we only randomly augment half of the images present in a mini-batch to preserve the information provided by the initial source domains. An example of style mixed images, along with an illustration of our proposed method, is provided in Fig. \ref{fig:cyclemix}. After augmentation, we pass the mixed images through a feature extractor and train a classifier to attain their final labels.

\section{Experiments}

\subsection{Experimental Setup}
To validate our approach, we use the publicly available and widely-adopted \textbf{PACS}\cite{Li_2017_ICCV} dataset. PACS contains images from 4 
separate style domains. As its name suggests, samples can either originate from the \textbf{P}hoto, \textbf{A}rt Painting, \textbf{C}artoon or \textbf{S}ketch domain and can be one of 7 classes. As in standard DG experimental setups, we follow the \textit{leave-one-domain-out} cross-validation protocol \cite{Li_2017_ICCV, gulrajani2021in}, meaning that in each training iteration we train a model on 3 source domains and evaluate on a single target domain. For our experiments, we use the DomainBed  \cite{gulrajani2021in} codebase and train a ResNet-50 feature extractor on a single NVIDIA A100 GPU card.

%% CLASSIC TABLE PACS
\begin{table}[H]\centering
	\begin{center}
		\caption{Top-1\% accuracy results on the \textbf{PACS} (left) and \textbf{VLCS} (right) datasets. The columns denote the target domains. The top results are highlighted in \textbf{bold} while the second best are \underline{underlined}.}
		\label{tab:pacs}
		\begin{tabular}{c|cccc|c}
			\toprule
			%		\hline
			\noalign{}%\smallskip}
		%		~ & \multicolumn{5}{c||}{\textit{PACS}} & \multicolumn{5}{c}{\textit{VLCS}} \\
		\textbf{Method} & \textbf{Art} & \textbf{Cartoon} & \textbf{Photo} & \textbf{Sketch} & \textbf{Avg} \\
		%\noalign{\smallskip}
		\midrule
		%		\hline
		%\noalign{\smallskip}
		%		\textit{ResNet-18} &&&&&\\
		%		ERM    
		%		& 79.68 & 77.76 & 88.91 & 75.72 & 80.53 \\
		%		
		%		CUTOUT    
		%		& 83.90 & 78.60 & 97.30 & 73.50 & 83.33  \\
		%		
		%		CUTMIX    
		%		& 83.90 & 78.60 & 97.30 & 73.50 & 83.33  \\
		%		
		%		MIXUP
		%		& 83.90 & 78.60 & 97.30 & 73.50 & 83.33  \\
		%		
		%		\midrule
		%		
		%		\textbf{CycleMix}
		%		& \textbf{}  & \textbf{} & \textbf{} &  & \textbf{} \\
		%		
		%		\midrule
		%\hline
		
		%		\textit{ResNet-50} &&&&&\\
		ERM    
		& 85.4 & 75.4 & 95.9 & 77.1 & \underline{84.8}  \\

		CUTOUT    
		& 81.8 & 81.8 & 95.8 & 78.1 & 84.2  \\
		
		CUTMIX    
		& 79.9 & 75.9 & 96.6 & 76.0 & 82.1  \\
		
		SagNet
		& 84.5 & 79.5 & 95.7 & 78.1 & 84,4 \\
		
		MIXUP
		& \underline{86.5} & 76.6 & \textbf{97.7} & 76.5 & 84.3  \\
		
		\midrule
		\textbf{CycleMix}
		& \textbf{87.7}	& \textbf{82.0} & \underline{96.6} & \textbf{79.9} & \textbf{86.6} \\
		\hline
		
	\end{tabular}
\end{center}
\end{table}

\subsection{Results}
To evaluate the effectiveness of CycleMIX as an augmentation approach, we 
rank its improvement against established techniques such as CUTOUT, CUTMIX and MIXUP. We also use SagNets as a baseline to demonstrate the efficacy of our method, along with a vanilla ResNet-50\footnote{Standard image augmentations are used during data loading for each method, such as random resized crop, random horizontal flip, color jitter and normalization.}. The results of our experiments are presented in Table \ref*{tab:pacs}.

It is apparent that the proposed method has a clear 
advantage over the baselines, as it surpasses the second best performing model by an average of around 2\%. With the exception of \textit{Photo}, \textit{CycleMix} yields the best results in every other target domain.

\section{Conclusion}
In this work we propose \textit{CycleMix}, a method aimed to alleviate the 
problems poised by style biased predictions in the DG setting. We argue that  mixing the different styles present in a convolutional neural network's training data, a model can be pushed to focus on the invariant features and extract robust representations. The above claim is supported by experimental results on PACS, a dataset containing images from 4 distinct style distributions, where our method surpasses previously proposed algorithms. However, one of the key limitations is that an increase in the number of source domains corresponds to an increased of trained CycleGANs, which can prove computationally infeasible. In future work, we intend to train StarGAN models which were proposed for multi-domain image-to-image translation and explore our methodology on additional datasets.

\begin{acks}
The work leading to these results has been funded by the European Union under Grant Agreement No. 101057821, project RELEVIUM. Views and opinions expressed are however those of the authors and do not necessarily reflect those of the European Union or the granting authority (HaDEA). Neither the European Union nor the granting authority can be held responsible for them.
\end{acks}

\bibliographystyle{ACM-Reference-Format}
%%\bibliography{sample-base}
\bibliography{biblio}

\end{document}